\documentclass[letterpaper, 10 pt, conference]{ieeeconf} 
 \usepackage{booktabs}

\IEEEoverridecommandlockouts

\usepackage{graphicx}
\usepackage{caption}
\usepackage{algpseudocode}
\usepackage{hyperref}
\usepackage{amsmath, amssymb}
\usepackage{fancyhdr}
\usepackage{lipsum} 

\title{\LARGE \bf
Labeling Sentences with Symbolic and Deictic Gestures via Semantic Similarity
}

\author{Ariel Gjaci, Carmine Tommaso Recchiuto, and Antonio Sgorbissa%
\thanks{The work took place at the University of Genova, Genova, Via Balbi 5, Italy. {\tt\small ariel.gjaci@edu.unige.it, carmine.recchiuto@dibris.unige.it, antonio.sgorbissa@unige.it}}%
}

\fancypagestyle{plain}{
    \fancyhf{}

    \fancyfoot[C]{\footnotesize
    \textit{© 2024 IEEE. Personal use of this material is permitted. Permission from IEEE must be obtained for all other uses, in any current or future media, including reprinting/republishing this material for advertising or promotional purposes, creating new collective works, for resale or redistribution to servers or lists, or reuse of any copyrighted component of this work in other works. This article has been accepted for publication in IEEE ROMAN 2024. The final published version is available at \textit{[DOI will be inserted here once available]}.
    }
    }
}

\begin{document}

\pagestyle{plain}

\maketitle
\thispagestyle{plain}

\begin{abstract}

Co-speech gesture generation on artificial agents has gained attention recently, mainly when it is based on data-driven models. However, end-to-end methods often fail to generate co-speech gestures related to semantics with specific forms, i.e., Symbolic and Deictic gestures. In this work, we identify which words in a sentence are contextually related to Symbolic and Deictic gestures. 
Firstly, we appropriately chose 12 gestures recognized by people from the Italian culture, which different humanoid robots can reproduce. 
Then, we implemented two rule-based algorithms to label sentences with Symbolic and Deictic gestures. The rules depend on the semantic similarity scores computed with the RoBerta model between sentences that heuristically represent gestures and sub-sentences inside an objective sentence that artificial agents have to pronounce. 
We also implemented a baseline algorithm that assigns gestures without computing similarity scores. 
Finally, to validate the results, we asked 30 persons to label a set of sentences with Deictic and Symbolic gestures through a Graphical User Interface (GUI), and we compared the labels with the ones produced by our algorithms. For this scope, we computed Average Precision (AP) and Intersection Over Union (IOU) scores, and we evaluated the Average Computational Time (ACT). Our results show that semantic similarity scores are useful for finding Symbolic and Deictic gestures in utterances.

\end{abstract}

\section{Introduction}
Co-speech gestures are crucial in communication between humans \cite{c1}. Indeed, when we speak, we unconsciously use body gestures to better express our intentions or to emphasize the verbal messages we want to convey. These gestures are also meaningful in the interaction between humans and artificial embodied agents \cite{c2,c3}, and between humans and social robots \cite{c4,c5,c6}. With recent advancements in machine learning, there has been an increased research focus on generating co-speech gestures.

According to \cite{c7,c8}, co-speech gestures can be classified into six categories: 1) Iconic gestures visually represent concepts we are verbally communicating e.g., flapping arms to represent a bird, 2) Metaphoric gestures represent abstract content, e.g., feelings, 3) Beat gestures emphasize the speech content and are often unrelated to semantics, 4) Deictic gestures are used to indicate objects, 5) Adaptors are self-touching movements, and finally, 6) Emblematic or Symbolic gestures carry specific meanings and are strongly related to culture, e.g., creating a circle with index and thumb to indicate ``OK".

Previous research on generating co-speech gestures for artificial agents has primarily focused on two approaches: 1) rule-based methods, where experts handcraft rules to map gestures to speech content, and 2) data-driven methods, where machine learning algorithms learn the mapping rules from data.
Rule-based methods produce smooth and human-like gestures but can only consider a limited set of gestures and mapping rules. 
Data-driven models overcome this limit by learning from data, but the generated gestures often lack natural forms and semantic dependency, particularly with Symbolic and Deictic gestures.
Moreover, the meaning of Symbolic and Deictic gestures can vary significantly depending on the culture of people using them \cite{c9,c10}. For instance, in India, the ``Namaste" gesture with palms pressed together is used for greeting, while in Italy, the same gesture is more commonly associated with praying.

To address these limitations, we propose two rule-based algorithms to identify and label the words within a given sentence that are semantically related to Symbolic and Deictic gestures. This approach can complement data-driven or other rule-based methods in hybrid configurations, allowing for the generation of various gesture types. For instance, data-driven methods can generate gestures for words not identified by the rule-based algorithms, while smooth transitions between methods can be achieved using interpolation techniques \cite{c11}.  

In this work, we focus only on Symbolic and Deictic gestures for three main reasons: 1) they depend on the speech semantics and the culture of people using them, 2) they have specific and easily recognizable shapes, 3) they are not frequently used, so they will not look repetitive when reproduced by a conversational agent.

The proposed algorithms are based on semantic similarity scores \cite{c12} computed with the Cross-Encoder RoBerta model \cite{c13}. The core idea is to measure the similarity between a set of \textit{reference sentences} designed by humans, which represent the contexts in which specific gestures are expected to be generated, and sub-sentences within an \textit{objective sentence} that the artificial agent has to pronounce while using co-speech gestures.  

\begin{figure}
    \centering
    \includegraphics[width=1\linewidth]{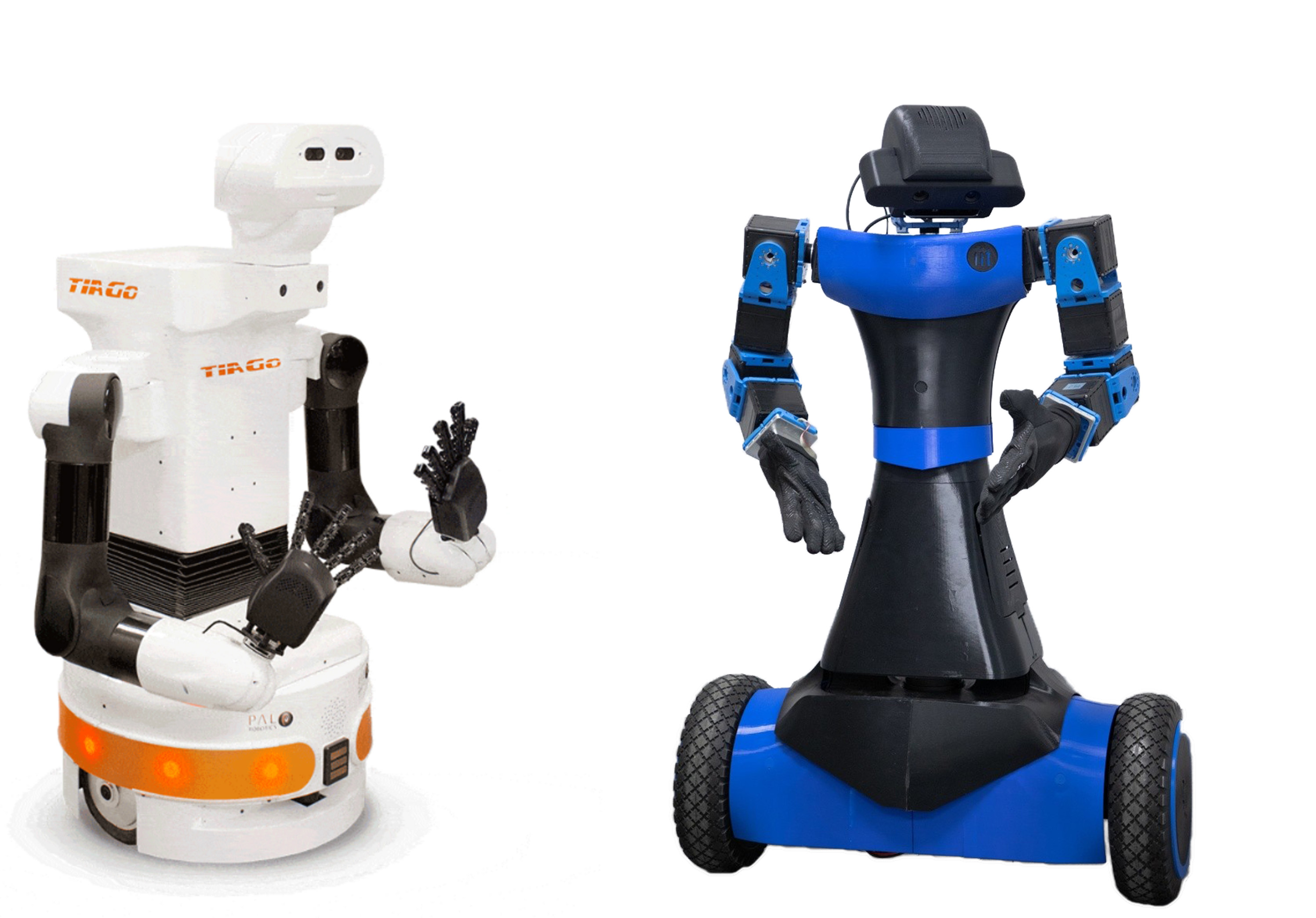}
    
    \caption{Tiago (on the left side) and Alter-ego (on the right side) robots.}    
        \label{figRob}
        \vspace{-10pt}

\end{figure}

We chose gestures from \cite{c10} and \cite{c14} recognizable in Italian culture and reproducible using only the upper body. This limitation ensures easier execution by social robots which can control hands and finger movements, such as Tiago \cite{c15} or Alter-ego \cite{c16}, as shown in Figure \ref{figRob}. We then associated these gestures with a set of reference sentences chosen by human experts.  
 
To test our approach, we asked 30 Italian participants to label two sets of 300 sentences generated using a ChatGPT language model. We compared their labels with those produced by our labeling algorithms by computing Intersection Over Union (IOU) and Average Precision (AP). Additionally, we measured the Average Computational Time (ACT) required by each algorithm to label a sentence. Note that the developed algorithms only require a set of reference sentences and a pre-trained LLM model, such as RoBerta, to function. The collected data serves solely as ground truth to validate our methods.

The paper is divided as follows: Section \ref{sec1} covers related works. Section \ref{sec2} details the methodology. Section \ref{sec3} describes the experimental setup. Section \ref{sec4} presents the results. Section \ref{sec5} provides conclusions and suggestions for future research.

 \section{Related work}
 
 \label{sec1}
 Co-speech gesture generation approaches can be divided into two main categories: \textit{rule-based} and \textit{data-driven}. 
 \subsection{Rule-based}
 Rule-based methods are commonly used in commercial robots because the generated gestures are defined through clear rules from human behavior studies. For instance, BEAT \cite{c17} analyzes input text and generates an appropriate gesture sequence based on rules from research on human conversational behavior.
In \cite{c18}, researchers created a library linking frequently used words to a set of handcrafted gestures. The approach in \cite{c19} involved generating gestures according to specific text styles, while \cite{c20} automatically identified and stored text-gesture mappings from videos without using learning models. At runtime, this method computes semantic similarity using GloVe embeddings \cite{c21} to find the most similar sequence of words in the learned text-gesture mapping and retrieve the corresponding gesture. However, the model required increasing memory with increasing mappings found from data, making it challenging to use in real-world applications. Additionally, the set of gestures considered is limited.

\subsection{Data-driven}
Data-driven methods aim to learn the mapping rules between gestures and speech features, allowing for the generation of a wide variety of gestures but often with less control over the learned mappings. These approaches can be further categorized into probabilistic, generative, and LLM-based.
 
 \subsubsection{Probabilistic} they rely on probabilistic models to map speech features to a set of animated gestures. For example, \cite{c22} estimated five expressive gesture parameters from speech-audio. At runtime, speech audio is used as input to generate a sequence of gestures by selecting, for each time frame, the gesture with the most similar parameters. 
 The work in \cite{c23} used morphemic analysis on utterances to predict the most likely gesture sequence. The utterance is first segmented into expression units using a Random Forest model, and then another model associates each expression unit with a gesture.
In \cite{c24}, a kNN-based algorithm retrieves gestures from an audio-gesture database using similarities computed on poses and audio features. The retrieved sequences are then refined using Generative Adversarial Networks (GANs) \cite{c25}. While this method overcomes the issue of limited gestures by using the generative network, its computational cost increases with the database size.

  \subsubsection{Generative} they use generative models like Transformer architectures \cite{c26} and Diffusion models \cite{c27} to overcome the problem of limited gestures. These methods learn end-to-end mappings from speech features to gestures,  generating then new gestures from speech input.
 For example, in \cite{c28}, the authors extracted speech audio and poses from TED Talks videos of Indian speakers and learned mappings between poses and audio using GANs.
Similarly, \cite{c29} extracted poses, audio, text, and speaker identity features from a TED Talks dataset and used an adversarial scheme to learn a mapping between gestures and the other features. 
 The work in \cite{c30} used a motion-capture dataset to extract gestures, text, gender, handedness, intended emotion, and acting tasks (narration or conversation). The researchers learned a mapping between text embeddings generated using Transformers and these features, excluding gestures. This combined representation, along with past gestures, was then fed into a Transformer decoder to generate gestures for subsequent time steps.
While these methods can generate a wide variety of gestures, they often lack control over the generation process, producing gestures with forms unrelated to speech semantics.

\subsubsection{LLM-based} they use Large Language Models (LLMs) to enhance the semantical dependency of generated gestures. For instance, \cite{c31} used ChatGPT to determine the gesture types and timings of a sentence, retrieved the corresponding gestures from a database, and then combined them with rhythmic gestures generated by generative models. 
Similarly, in \cite{c32}, authors used ChatGPT prompts to retrieve gesture types for sentences from annotated examples and to suggest gesture types for sentences without annotated examples.
While both \cite{c31} and \cite{c32} use LLMs as black boxes to predict gesture types, our approach employs heuristics on semantic similarity scores computed by the RoBerta model \cite{c33} to label only specific words related to a set of predefined gestures. 
This method allows for more precise labeling, greater control over the LLMs' behavior, and integration with other methods more suitable for generating different types of gestures.

\section{Methodology}
 \label{sec2}
In this study, we propose two rule-based \textit{labeling algorithms} designed to annotate sentences with Symbolic and Deictic gestures based on semantic similarity scores produced by a recent Large Language Model (LLM), RoBerta. Additionally, we introduce a baseline \textit{labeling algorithm} that uses a statistical approach to label sentences. 
LLMs have demonstrated their ability to capture text semantics by achieving high scores in the Semantic Textual Similarity (STS) benchmark \cite{c12}. This suggests that it is possible to associate a semantic-dependent gesture with a set of reference sentences that embed the contexts in which the gesture is often reproduced. 
The choice of gestures was informed by studies \cite{c7}, \cite{c10}, \cite{c14}, while the reference sentences were selected by four experts in human-robot interaction.

More formally, we defined a set of sets of sentences ${D = \{S_1,..., S_l,...S_k\}} $, i.e. the \textit{reference sentences}, that we heuristically assume are capable of capturing the contexts in which $k$ gestures ${G = \{g_1,...,g_l,...,g_k\}}$ are generated. Each ${S_l = \{s_1,...,s_i,...,s_{n_l}\}}$ is a subset of $D$ that contains a set of $n_i$ sentences $s_i$ representing the contexts on which the corresponding gesture $g_l$ is supposed to be generated. 

Given an \textit{objective sentence} $S_{obj}$ composed of a sequence of words ${W_{obj}=\langle w_1,...,w_{n_{obj}}\rangle}$, we aim to find all sequences of words \begin{math} {W_g \hookrightarrow W_{obj}} \end{math}
 related to the semantics of the gestures $G$. Each sequence inside ${W_g}$ is manually associated with a corresponding gesture by a human participant and called \textit{real label}, while ${W_p}$ contains the labels predicted by our algorithms. The number of words inside a label will be called \textit{window size}.

\begin{figure} 
    \centering
    \includegraphics[width=1\linewidth]{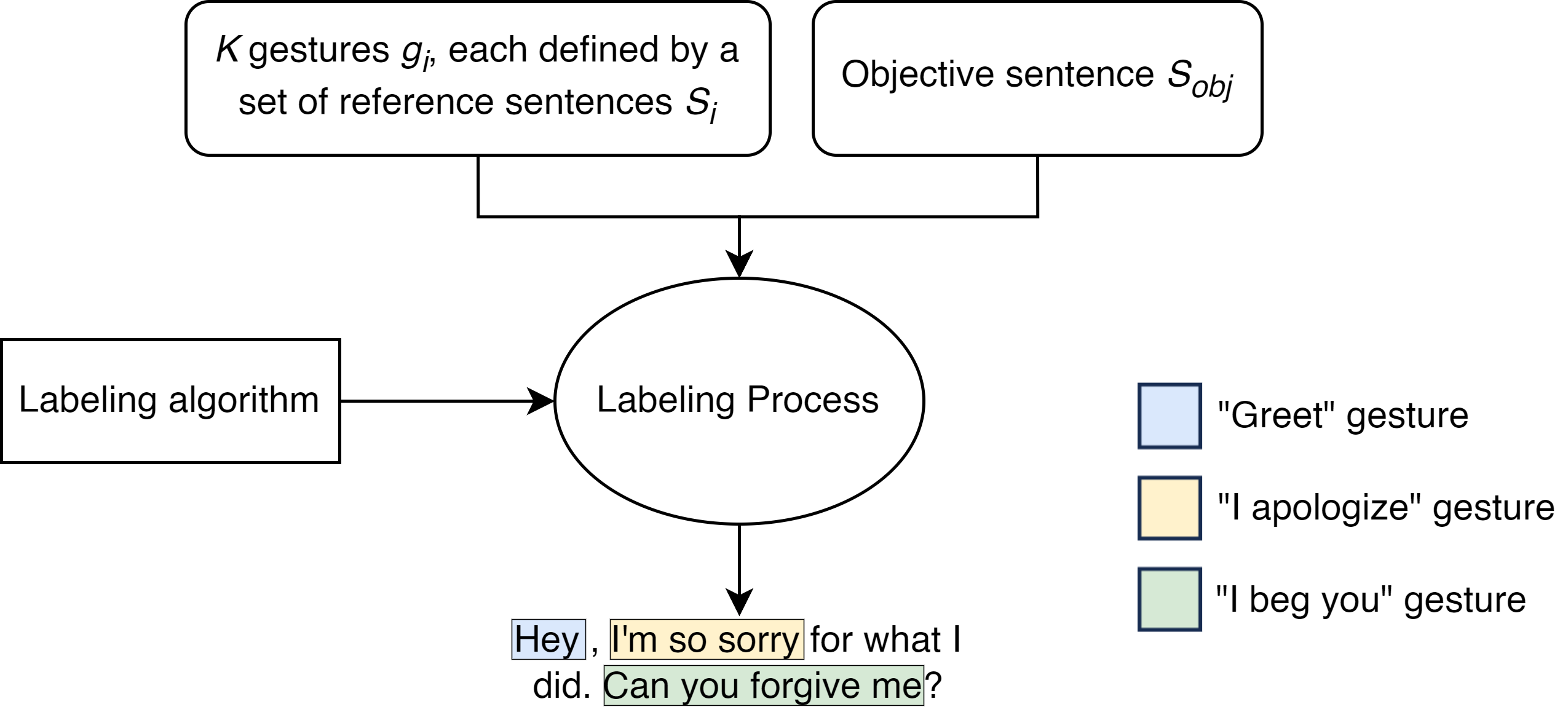} 
    
    \caption{Diagram representing the approach. Given a set of $K$ gestures $g_i$, each represented by a set of \textit{reference sentences} $S_{i}$ and an \textit{objective sentence} $S_{obj}$, a labeling algorithm is used to produce labels for $S_{obj}$. In the given sentence, three labels are produced for three different sequences of words, forming 
      $W_p = \langle\langle\textit{Hey}\rangle_{\textit{Greet}},  \langle\textit{I'm, so, sorry}\rangle_{\textit{I apologize}}, \\ \langle \textit{Can, you, forgive, me}\rangle_{\textit{I beg you}}\rangle$. }     

    \label{fig1}
    \vspace{-10pt}

\end{figure}

\begin{figure}[htbp]
    \centering
    \includegraphics[width=1\linewidth]{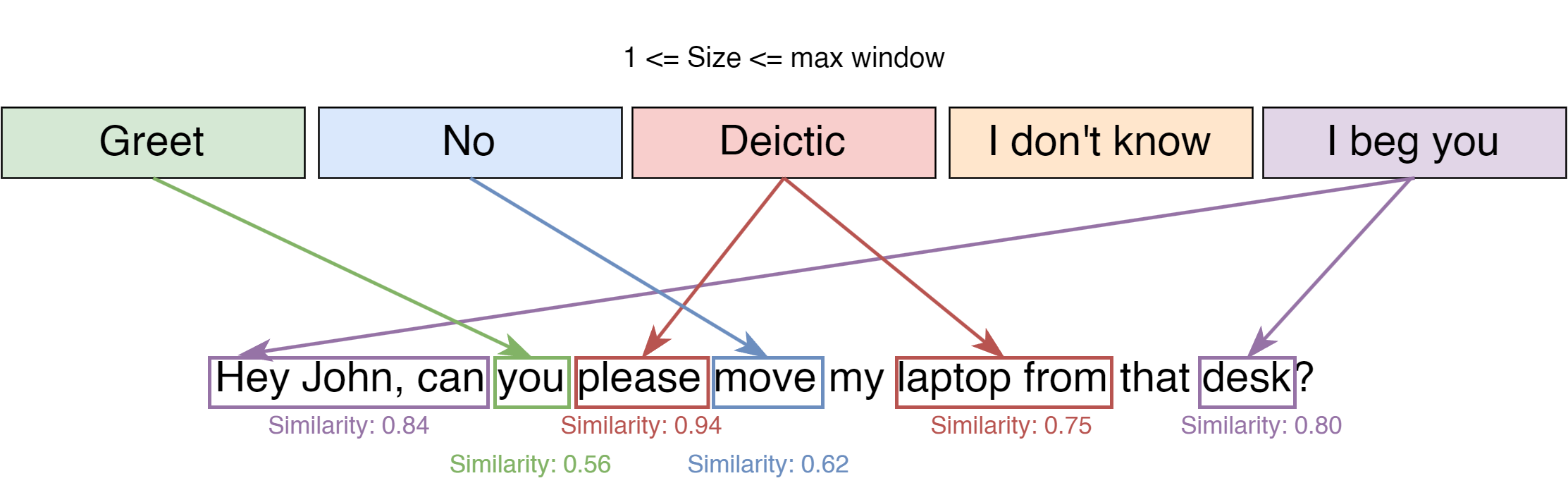}
    
    \caption{Practical example of the Baseline algorithm. Labels are not assigned depending on similarity scores but depending on a predefined label distribution.}

    \label{fig2}
    \vspace{-10pt}

\end{figure}

\begin{figure}[htbp]
    \centering
    \includegraphics[width=1\linewidth]{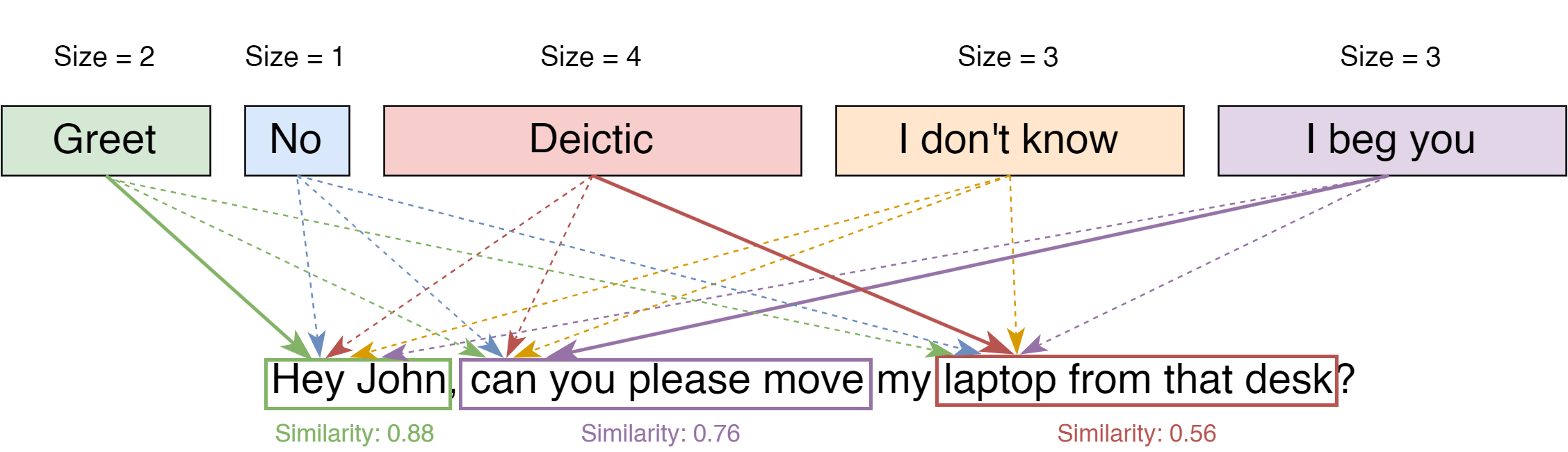} 
    
    \caption{Practical example of the Fixed Window algorithm. Window sizes have a fixed value for each gesture. All similarities are computed starting from the words pointed by arrows. The non-labeled words have a similarity score less than $th_0$ for all the gestures, so they are skipped.}

    \label{fig3}
    \vspace{-12pt}

\end{figure}

\begin{figure}[htbp]
\vspace{-10pt}
    \centering
    \includegraphics[width=1\linewidth]{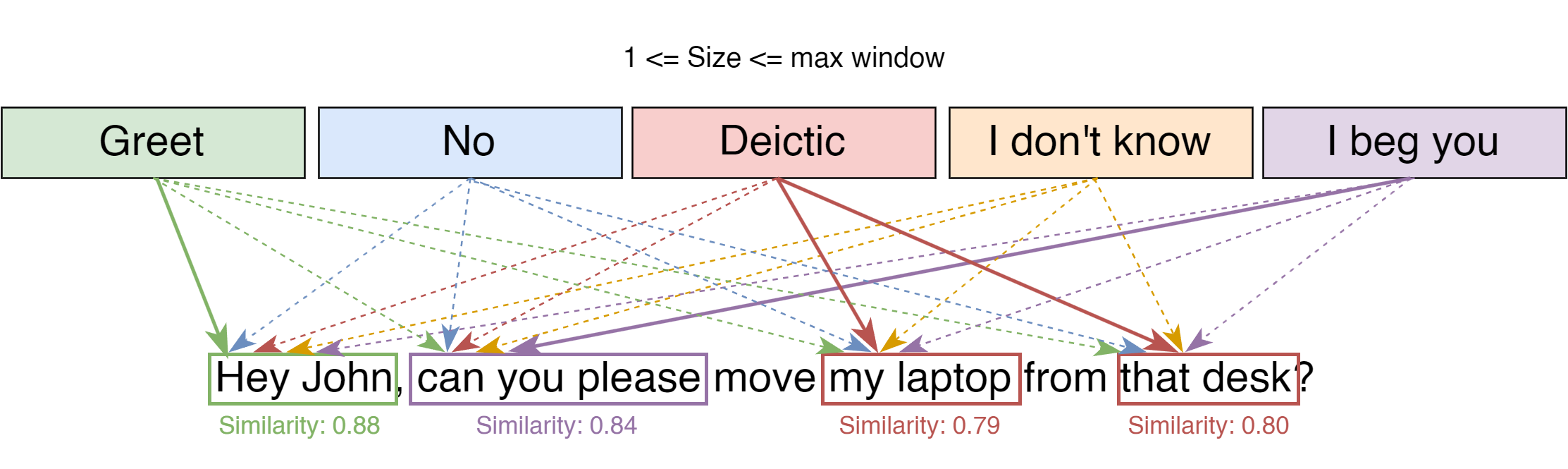} 
        \caption{Practical example of the Moving Window algorithm. The labeling process is similar to the Fixed Window algorithm, however, this time window sizes are not fixed and can have any value between $1$ and $w_{max}$.}   

    \label{fig4}
    \vspace{-5pt}

\end{figure}

For example, ${W_g = \emptyset}$ if no words inside $S_{obj}$ are related to the semantics of $G$. If, instead, $S_{obj}$ contains a subsequence of three words ${\langle w_1,w_2,w_3 \rangle}$ related to $g_1$, and another subsequence of two words ${\langle w_5,w_6 \rangle}$ related to $g_3$, then ${W_g = \langle {\langle w_1,w_2,w_3 \rangle}_{g_1},{\langle w_5,w_6 \rangle}_{g_3}\rangle}$. Ideally, our algorithms should predict ${W_p \equiv W_g}$. This approach is illustrated in Figure \ref{fig1} with a practical example.

In this work, we assume that each word is assigned to at most one gesture. We also assume that each of the Symbolic and Deictic gestures we chose is used in different contexts from the others, so ${s_i  \neq s_j \; \; \forall i \neq j}$, given that ${s_i}$ represent a specific context through semantics.

Identifying $D$ and $W_g$ is challenging since co-speech gestures are idiosyncratic \cite{c7}, leading different people to use different gestures also when context and pronounced words are the same. However, Symbolic and Deictic gestures are culturally and semantically encoded \cite{c7},\cite{c9}, so we assume that people from the same culture will use similar gestures in similar contexts. 
In the following, we describe the algorithms we developed.

\subsection{Baseline algorithm}
The Baseline algorithm uses a statistical approach to label a sentence $S_{obj}$. If ground truth labeled sentences are provided by human participants, the algorithm labels using the following statistics for each gesture:

\begin{equation*}
P(\textit{label with gesture}_i \,|\, \textit{gt\_labels}) =  \frac{\textit{total labels gesture}_i}{\textit{total labels }} 
\end{equation*}  

\begin{equation*}
win_{g_l} | \, \textit{gt\_labels} \sim \mathcal{N}(\mu_{g_l}, \sigma_{g_l}^2)
\end{equation*} 
where \textit{gt\_labels} represents the ground truth labels, $\mathcal{N}$ represents a normal distribution, while $\mu_{g_l}$ and $\sigma_{g_l}^2$ are the mean and the variance of the window size for the gesture $g_l$, indicated as $win_{g_l}$, extracted from ground truth labels. Note that the window size $win_{g_l}$ is rounded to the nearest integer since it cannot have floating values. We also limited the maximum value of $win_{g_l}$ to a constant $w_{\max}$, which is the same also for the following algorithms. Our labeling choice enables the labeling of sentences based on the ground truth distribution. In the absence of ground truth data, alternative distributions can be utilized.

We finally assign to each label a similarity score with a random value above a certain threshold $th_0$, the same used for the following algorithms. A practical example of this algorithm is shown in Figure \ref{fig2}. 

\subsection{Fixed Window algorithm }
The Fixed Window algorithm uses a fixed window size for each gesture $g_l$, with ${l = 1,...,k}$. This means that every time we compute the semantic similarity between a sequence of words inside a sentence and $s_i \in S_l$, the window size depends on the $g_l$ related to $S_l$, and this dependency is indicated with $win_{g_l}$ as previously shown. For example, if $g_1$ is assigned a window size $win_{g_1} = 3$, then ${\forall s_i \in S_1}$, semantic similarity is computed as 
${SemSim(s_i,S_{obj}[w_j,...,w_{j+3}])}$.
Given $S_{obj}$, starting from the word having index $j=1$,  
we compute ${v_i = SemSim(s_i, S_{obj}[w_j,...,w_{j+win_{g_l}}])}$ 
for every ${i=1,...,n_l}$, with ${s_i \in S_l}$, and for every ${l = 1,...,k}$. 

At the end, we have a set $Z$ with $l$ subsets ${V_l \in Z}$, each subset ${V_l}$ containing $n_l$ values ${v_i \in V_l}$. We then find: 
\begin{equation*}
    \begin{array}{c}
        v^* = \max\biggl\{\{v_i \in V_l : i = 1,...,n_l\} 
        \biggm| V_l \in Z, l = 1,...,k\biggr\}   
    \end{array}
\end{equation*}
that is the highest value among all the $v_i$ computed. If $v^*$ is not above a given threshold $th_0$, we update $j$ to $j+1$ (i.e., we move the windows forward), and restart the computations. 
Otherwise, we extract the sentence $s^*$, the gesture $g^*$, and the window size $win_{g^*}$ that yielded $v^*$. 
We finally assign the label $g^*$ to the words ${\langle w_j,...,w_{j+win_{g^*}} \rangle}_{g^*}$, add the label to $W_p$, and update the index $j$ to ${j+win_{g^*}}$. 
The set of all $v^*$ computed will be called $V^*$ in the next.
If ${j+win_{g_l}}$ exceeds the sentence length $n_{obj}$, $g_l$ is skipped and not considered for the computation of $v^*$. We continue these steps until we reach the end of the sentence.
 
To determine the window size for each gesture, we run this algorithm over two sets of 300 sentences produced with an OpenAI language model, repeating the process for all possible window sizes (${win = 1,...,w_{max}}$) and gestures $G$. We associate each gesture with the window size that returns the maximum ${\overline{V}^* - \sigma_{V^*}}$, where ${\overline{V}^*}$ and ${\sigma_{V^*}}$ are the average and the standard deviation of all the $v*$ values computed across all the sentences. We considered as valid only the window sizes that obtained at least 10 ${v^*}$ values, which must be above the threshold $th_0$.

While the computational cost of finding the best window size increases with the number of gestures and sentences, this process can be computed offline. This significantly reduces the computations needed to label $S_{obj}$ and makes the method more suitable for real-time applications. It is important to note that data is not strictly required: if a set of sentences is provided, then the algorithm computes the best window size for each gesture by using statistics and similarity scores, otherwise values can be associated randomly or given according to preferences. 
A practical example of the Fixed window is shown in Figure \ref{fig3}.
 
\subsection{Moving Window algorithm}
The Moving Window algorithm does not use a fixed window size for each gesture $g_l$, with ${l = 1,...,k}$. For this reason, we also aim to find $win^*$ that yields the maximum value $v^*$. Given a sentence $S_{obj}$ with $n_{obj}$ words, starting from index $j=1$, ${j=1...n_{obj}}$, we compute ${v_{i,win} = SemSim(s_i,S_{obj}[w_j,...,w_{j+win}])}$ with $s_i\in S_l$, for each ${i=1,...,n_l}$, ${win = 1,...,w_{max}}$ , and for every ${l = 1,...,k}$. 
We then compute:
\begin{equation}
    \begin{array}{c}
    v^* = \max\biggl\{\{v_{i,win} \in V_l : 
    i = 1,...,n_l\, , \\[1mm]  win = 1,...,w_{max}\} \nonumber 
        \biggm| V_l \in W, l = 1,...,k\biggr\}
    \end{array}
\end{equation}
If $v^*$ is not above a given threshold $th_0$, we update $j$ to $j+1$ and repeat the operations above.  
Labeling sentences using this method is computationally heavy and may not be suitable for real-time applications. Here, we computed all the possible semantic similarity scores in advance to avoid further complications.

Given that we use a variable window size, we added two controls after computing $v^*$. First, we check if the gesture $g^*$ associated with $v^*$ is consistent by considering additional words in $S_{obj}$. 
This check is performed by calculating, when it is possible, ${v_{check} = SemSim(s^*, S_{obj}[w_j,...,w_{j+win^*+1}])}$, where $s^*$ and $win^*$ are respectively the window size and sentence related to the context of $g^*$ that returned $v^*$. If ${v_{check}-v^*>th_1}$, where $th_1$ is a threshold chosen heuristically by us, we do not accept the result and repeat the computation of $v^*$ for different gestures, excluding $g^*$. In a few words, we try to expand the moving window from $win^*$ to $win^*+1$ to check if the context changes. 
This process is repeated $p$ times, where $p$ is a hyperparameter: if we don't find a valid $v^*$ for $p$ tries, we update $j$ to $j=j+1$ and move the window forward. 

To show the necessity of this process, consider the example where $v^* = 0.9$, $s^* =$ \textit{"I love"}, $S_{obj}$ = \textit{``I love fighting"}. Even if $v^*$ is high, using a gesture related to \textit{``love"}, such as forming a heart shape with fingers, may not be the best choice. Indeed, by increasing the window size by one, it becomes clear that a gesture to represent \textit{``fight"}, like forming punches in front of the chest, would be more appropriate. In this case, we expect that $v_{check}$ computed with words ${\langle w_j,...,w_{j+win^*+1} \rangle}$ and $s^*$ would return a much lower value with respect to $v^*$, indicating a context change. Secondly, we check if there is another index $j_{check}$ between $j$ and ${j+win^*}$ that returns a better result ${v^*_{check} > v^*}$, with $s^*_{check}$, $win^*_{check}$ relative to ${v^*_{check}}$, and that also meets the first check. 
If such an index is found, we replace the result previously computed with the newer one, and we update $j$ to ${j_{check} + win^*_{check}}$. Otherwise, we take the previous result and update $j$ to ${j + win^*}$. 
The process is repeated until we reach the end of the sentence. 

While the computational cost of this algorithm is higher than the others, as it increases with the number of reference sentences and with the length of the processed $S_{obj}$, it should be noted that each semantic similarity score can be computed independently before finding $v^*$, making the algorithm highly parallelizable.
A practical example of the Moving Window algorithm is shown in Figure \ref{fig4}.

All three algorithms use a maximum window size of 10 words $w_{max}=10$. This limitation avoids capturing long contexts that are unlikely to represent the gestures. Additionally, we did not address the problem of synchronizing gestures with robot speech and movements, as we did not generate them in real-world scenarios. This problem may require further theoretical considerations, e.g., maintaining or repeating the gesture if its execution time is faster than the time needed to pronounce the labeled words \cite{c7}.

\begin{table*}[htbp]
    \centering
    \includegraphics[width=\linewidth]{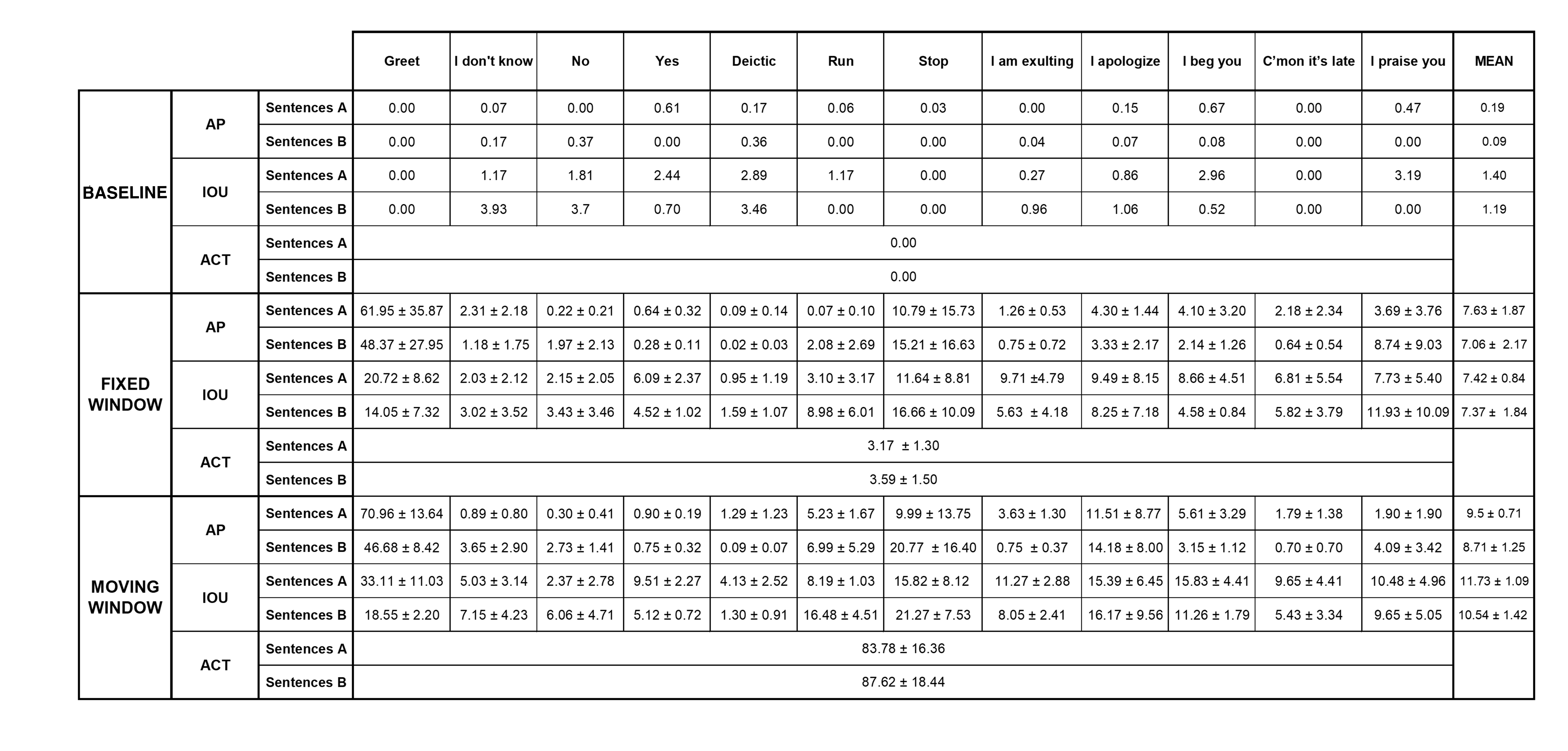}
    \vspace{-13pt}
    \caption{AP, IOU, and ACT scores obtained by running the described three algorithms on the two different sets of sentences. In this setup, $th_0 = 0.3$.
    All values are shown with average and standard deviation (being the algorithm executed multiple times with the different $D$ of the four experts), except for the Baseline algorithm which does not depend on predefined sentences. In the last column, we also show the Mean IOU and Mean AP. The values of AP and IOU are measured in percentages, with the maximum being 100, while ACT is measured in seconds.
}
    \label{tab:table1}
\end{table*}

\begin{table*}
    \centering
    \includegraphics[width=\linewidth]{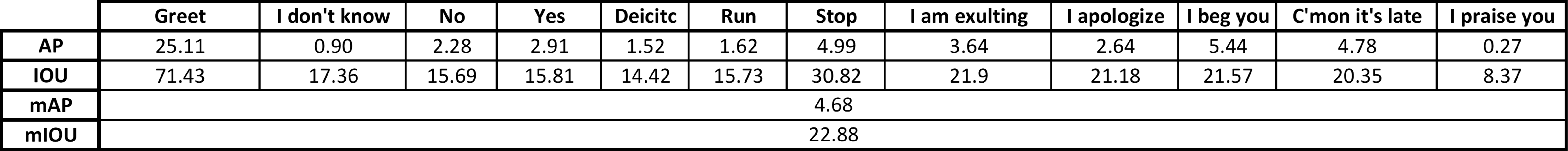}
    \caption{AP, IOU, and ACT scores obtained
    by comparing sentences labeled multiple times by different participants. This group of sentences is formed by 150 \textit{Sentences A} labeled two times each, and 150 \textit{Sentences B} labeled two times each.
}
    \vspace{-10pt}
    \label{tab:table2}
\end{table*}

\section{Experimental setup}
 \label{sec3}

 \begin{figure}
    \centering
    \includegraphics[width=1\linewidth]{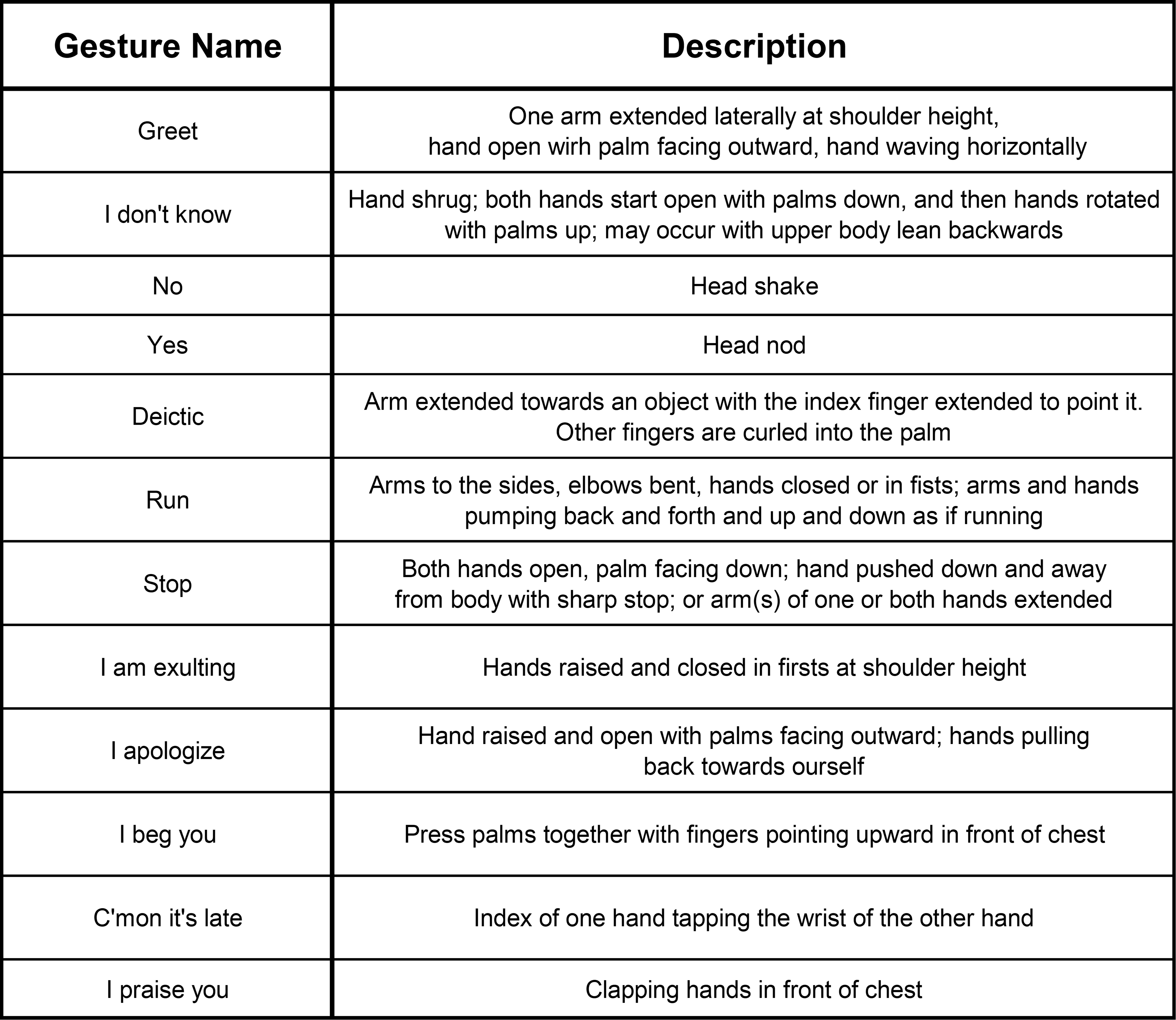} 
    \caption{Gestures considered in the experiments along with their brief description.}  
    \label{fig5}
    \vspace{-12pt}

\end{figure}

 \begin{figure}
    \centering
    \includegraphics[width=1\linewidth]{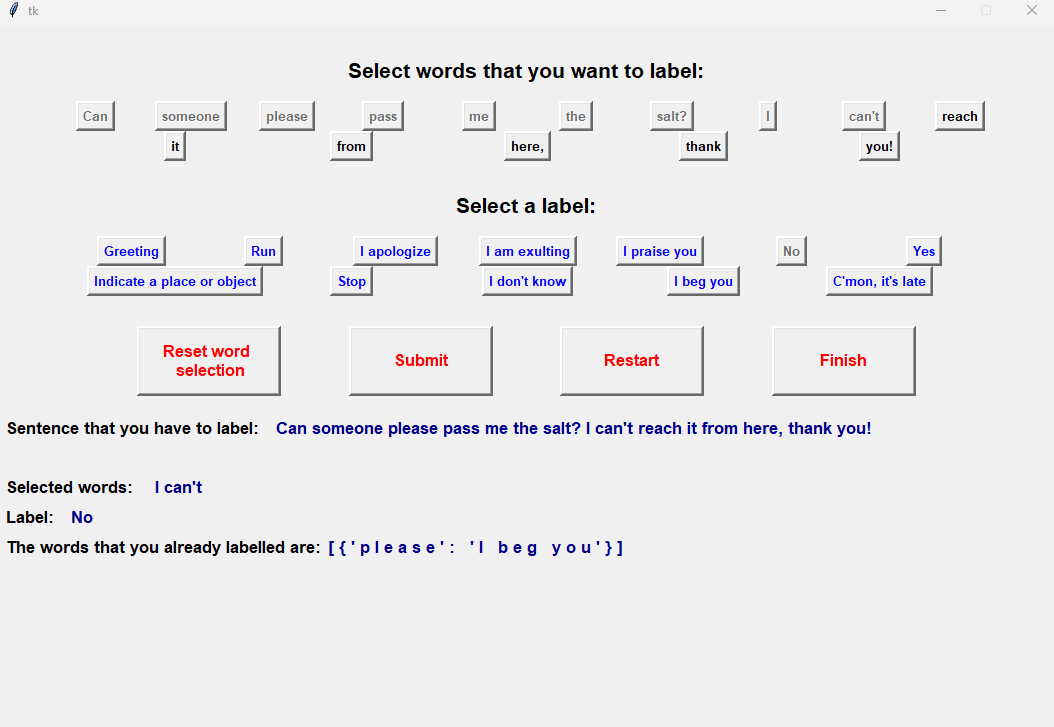}
    
    \caption{GUI used to label sentences. The considered sentence is ``Can someone please pass me the salt? I can't reach it from here, thank you!". One label is already produced, i.e. the word ``please" is associated with the gesture ``I beg you". Labels must be given sequentially by following the order of the words in the sentence. The words ``I can't" are currently selected to be associated with the label ``No", so all the words before ``can't" are no more available for selection.}
    \label{fig6}
    \vspace{-15pt}

\end{figure}

We tested the three algorithms by considering 11 Symbolic gestures from \cite{c10},\cite{c14}, and one Deictic gesture \footnote{Code and data are available on the following repository: \url{https://github.com/arielgj95/Gestures_labeling.git}}.

We selected all the gestures common to different cultures from \cite{c10} and gestures typical of the Italian culture from \cite{c14}. We then carefully picked a subset of gestures that can be easily reproduced by various social robots with movable hands and fingers, such as Tiago or Alter-ego. 

We recruited three Italian participants to validate the gestures before the experiments. We showed them all the gestures and the contexts in which they are used, and asked if they recognized and often used the gestures in the described contexts.  
Finally, we retained only the 12 gestures that received unanimous positive feedback and named them according to the conventions of the reference works \cite{c10} and \cite{c14}. The set of the chosen gestures is briefly described in Figure \ref{fig5}. It is composed by: \textit{Greeting}, \textit{I don't know}, \textit{No}, \textit{Yes}, \textit{Run}, \textit{Stop} \cite{c10}, \textit{I am exulting}, \textit{I apologize}, \textit{I beg you}, \textit{C’mon it’s late}, \textit{I praise you}, \textit{I am exulting} \cite{c14}, and a Deictic pointing gesture \cite{c7}.
Note that in this experiment we are only interested in the pointing gesture's relation to speech context, regardless of its direction. 
For each gesture, we also asked four experts on human-robot interaction to write four different sentences that reproduce the contexts in which the gestures are generated, thus obtaining four sets of reference sentences $D$. These sentences are needed to compute the semantic similarity scores with sub-sentences inside objective sentences. 

To produce a set of objective sentences $S_{obj}$ we adopted the following approach.
We generated two sets of 300 first-person sentences each by creating specific prompts for the OpenAI's ``gpt-3.5-turbo-16k" model. The first set, \textit{Sentences A}, was generated using a single prompt that required the model to produce 300 different sentences, each containing approximately 15 words and one or more contexts related to the considered gestures in various scenarios (school, cinema, etc.). The sentences were designed to be easy to understand, even by non-native speakers. The second set, \textit{Sentences B}, was  generated similarly but used multiple prompts, each related to a specific scenario (work, school, hobbies, etc.). For this set, the model was asked to generate sentences that may or may not include words related to gestures, making the task more challenging and better representative of real-world utterances. 
We ensured that each sentence was unique. %

We then recruited 30 native Italian speakers with at least B1 English proficiency to label \textit{Sentences A} and \textit{Sentences B}. Participants used a Graphical User Interface (GUI) we created for the task, shown in Figure \ref{fig6}, and labeled 30 sentences randomly picked from one set. 
 Participants were informed that each sentence could contain no gesture, one gesture, or multiple gestures, and were asked to imagine using appropriate gestures while speaking the sentence. 
 Finally, we asked participants to label words even if the Symbolic or Deictic gestures are not the most representative of the chosen sequence of words.  
While displaying the GUI, another monitor showed a \textit{gif} iteratively displaying the gestures and their names, along with Google Translate for translating unknown words if needed.
We decided to use Google Translate since the generated sentences were in English to better work with RoBerta, while all the participants we chose were native Italian speakers. In practice, participants did not use the translation tool, suggesting that ChatGPT-generated sentences were easy to understand even by non-native speakers.
We assume that using English instead of Italian does not significantly affect the results, given that Symbolic and Deictic gestures are strongly related to speech context, which is not language-dependent.
We aim to verify this assumption in future work by repeating the experiments with the native language spoken by participants. 

We finally ran the three algorithms by using only the reference sentences $D$ defined by one expert each time. For this process, we utilized the ``base" version of the Cross-Encoder RoBerta model, which we chose over Bi-encoder models \cite{c33} due to its superior performance on sentence similarity tasks. When testing the algorithms, the ``large" version of RoBerta was the best-performing pre-trained model on the STS benchmark, with the ``base" version showing similar performance while being significantly smaller and faster in inference. After running the algorithms, we compared the Average Precision (AP), Intersection Over Union (IOU), and Average Computational Time (ACT) for each gesture and each algorithm. 
These metrics are defined as follows: 
\[
 AP = \sum_{i=1}^n (R(t_i) - R(t_{i-1})) P(t_i) 
\] 
where $t_i$ is the $i$-th prediction threshold, $R(t_i)$ is the recall at threshold $t_i$, $P(t_i)$ is the precision at threshold $t_i$, and $R(t_{i-1})$ is the recall at the previous threshold $t_{i-1}$,
\[
IOU = \frac{\text{Area of Overlap}}{\text{Area of Union}}
\]
\[
ACT = \frac{\sum_{i=1}^{N} T_i}{N}
\] where \( T_i \) is the time taken to process the \( i \)-th sentence, and $N$ is the number of processed sentences.
We ran all the algorithms on the same PC under identical conditions. We repeated the previous steps three times by selecting each time a different $th_0$ value among $\{0.3, 0.6, 0.9\}$. We also tested how the algorithms perform when only one sentence for each gesture is randomly selected for each expert. To compute AP, we set the minimum IOU threshold to consider a prediction valid to $0.5$. For computing the total IOU for each gesture, we set the minimum acceptable similarity score for each predicted label to $0.5$.

\section{Results}
 \label{sec4}

We computed AP (in \%), IOU (in \%), and ACT (in seconds) for all the possible $th_0$ values in the set $\{0.3, 0.6, 0.9\}$, using the four $D$ provided by experts, the two sets of sentences generated by exploiting an OpenAI model, and the ``base" version of the Cross-Encoder RoBerta model. We show in Table \ref{tab:table1} the AP, IOU, Mean AP, Mean IOU, and ACT values obtained by averaging the results from the four experts' with $th_0 = 0.3$. Mean AP and Mean IOU, represented in the last column of the table, are defined as the average AP and IOU, respectively, over all the possible gestures, averaged then again by considering the results of the four $D$ sets. 

From Table \ref{tab:table1}, we observe significant differences in ACT values among the three algorithms: the Baseline algorithm labels sentences almost instantly, the Fixed Window algorithm requires about $3$s, while the Moving Window algorithm needs more than $80$s for both \textit{Sentences A} and \textit{Sentences B}. The higher computational cost of the Moving Window algorithm is due to the need to compute all possible semantic similarity scores before labeling, whereas the Fixed Window algorithm computes only the necessary scores at runtime. The Baseline algorithm, instead, assigns labels with a statistical approach, with similarities scores assigned randomly, without requiring any computation by RoBerta. By comparing the Mean IOU and Mean AP of the Baseline algorithm with the others, we show that semantic information is meaningful to identify when Symbolic and Deictic gestures may be generated. 
We also obtained higher Mean IOU and Mean AP values when labeling \textit{Sentences A} instead of \textit{Sentences B} with the Moving Window and the Fixed Window algorithms, although AP and IOU for individual gestures are not always higher. 
There is also a noticeable variability in AP and IOU scores related to different gestures using the same algorithm. For example, the \textit{Greet} gesture achieved $70.96$ AP between ground truth and predicted labels of \textit{Sentences A} when using the Moving Window algorithm, while the \textit{No} gesture obtained only $0.30$. This discrepancy is not due to the different label frequency, as volunteers labeled \textit{No} and \textit{Greet} a similar amount of times (45 vs. 42). This suggests that there is a significant variability and possible ambiguity in how people associate Symbolic and Deictic gestures with contexts, even when they belong to the same culture. To verify this fact, we combined all \textit{Sentences A} and \textit{Sentences B} labeled multiple times by different participants, and recomputed AP and IOU scores to check for consistency.
The results, shown in Table \ref{tab:table2},  are similar to those obtained with the Fixed Window and Moving Window algorithms. Although AP and IOU scores are generally slightly higher, their value remain low, especially when compared to \textit{Greet} gesture, which achieved the highest IOU ($71.43$) and AP ($25.11$) scores. The Mean IOU was $22.88$ while Mean AP was $4.68$. This outcome support the need for additional metrics to better validate our findings. 

Increasing $th_0$ to $0.6$ slightly worsened the results. For instance, using the Fixed Window algorithm with \textit{Sentences A}, Mean AP dropped to $6.32$, Mean IOU to $5.36$, and ACT increased to $6.14$s. With the Moving Window algorithm and \textit{Sentences A}, Mean AP decreased to $8.57$, Mean IOU to $9.92$, while ACT remained the same due to the need to compute all the possible semantic similarity scores. The Fixed Window algorithm's increasing ACT and decreasing Mean AP and Mean IOU values indicate that it often fails to find labels with a similarity score greater than $th_0$, leading to do more computations to label the sentence. This is confirmed by evaluating the same algorithm with $th_0 = 0.9$: all scores decreased except ACT which became $7.32$s.

Finally, when selecting a random sentence to represent the gesture's context, both algorithms showed a decrease in ACT values. For \textit{Sentences A} with the Moving Window algorithm and $th_0 = 0.3$, ACT reduced from an average of $83.78$s to $20.84$s. With the Fixed Window algorithm, ACT reduced to $1.06$s. However, the Fixed Window algorithm also led to a significant decrease in Mean AP, going from $7.63$ to $0.82$, and Mean IOU, going from $7.42$ to $3.57$. The Moving Window algorithm showed less significant drop, with Mean AP decreasing from $9.5$ to $7.29$ and Mean IOU from $11.73$ to $6.49$. This indicates that the deeper analysis performed by the Moving Window algorithm allows it to better recognize similarities, even when using a single sentence to represent the context in which a gesture is generated.

 \section{Conclusions}
 \label{sec5}

 In this paper, we presented a rule-based approach leveraging semantic similarity scores to link Symbolic and Deictic gestures with word sequences inside objective sentences. We developed three algorithms: a baseline that labels sentences without semantic consideration, one that uses precomputed fixed windows for fast labeling, and another that tries different windows to better identify semantic dependency. These algorithms require no training data, can be easily controlled and expanded with additional gestures, integrated with other methods, and are suitable for parallel processing to meet real-time constraints.

Our approach has also several limitations. Firstly, while the comparison with ground truth labels revealed that semantics plays a key role in identifying Symbolic and Deictic gestures, the association between speech context and these gestures varies significantly across different gestures. This suggests the need to consider additional factors for more accurate identification. Secondly, although AP and IOU metrics provide some performance insights, they do not fully validate the quality of the produced labels.
The complexity of the considered problem and the idiosyncratic nature of gestures make AP and IOU values highly variable, even when comparing labels produced by people of the same culture. We plan to explore better subjective and objective metrics in future work.
Additionally, our methods rely heavily on the reference sentences provided by the designer, assuming complete and non-overlapping context representation. More advanced LLMs and different rule strategies may enhance performances and better manage ambiguities. Scalability is another challenge; our current methods are suited for a limited number of gestures and reference sentences. In future work, we aim to use Bi-encoders to store embeddings of reference sentences and commonly used sentences for better scaling, and eventually compare the results.

Finally, we plan to test our algorithms with other cultures for better validation and implement them within hybrid rule-based and data-driven architectures to generate a wide variety of gestures in artificial agents. We will eventually measure performance using more commonly used metrics to facilitate comparison with other state-of-the-art works.

\section*{ACKNOWLEDGMENT}
This work was carried out within the framework of the project "RAISE - Robotics and AI for Socio-economic Empowerment” Spoke 2 and has been supported by European Union – NextGenerationEU.

\end{document}